\documentclass[letterpaper, 10 pt, conference]{ieeeconf}  

\IEEEoverridecommandlockouts                              

\usepackage{graphics} 
\usepackage{epsfig} 
\usepackage{mathptmx} 
\usepackage{times} 
\usepackage{amsmath} 
\usepackage{amssymb}  
\usepackage{bm}

\usepackage{graphicx}
\usepackage{epstopdf} 
\usepackage{mathtools}
\usepackage{float}

\title{\LARGE \bf
STDepthFormer: Predicting Spatio-temporal Depth from Video \\ with a Self-supervised Transformer Model

}

\author{Houssem eddine BOULAHBAL$^{1,2}$, Adrian VOICILA$^{2}$ and Andrew I. COMPORT$^{1}$
\thanks{*This work was performed using HPC resources from GENCI-IDRIS (Grant 2021-011011931). The authors would like to acknowledge the Association Nationale Recherche Technologie (ANRT) for CIFRE funding (n°2019/1649).}
\thanks{$^{1}$CNRS-I3S, Côte d’Azur University,
        2000 Route des Lucioles, Sophia Antipolis, France
        }
\thanks{$^{2}$Renault Software Factory,
        2600 Rte des Crêtes, 06560 Valbonne, France 
        }%
}
\newcommand{\mbf}[1]{{\mathbf{#1}}}

\newcommand{\refsec}[1]{Sec.~\ref{sec:#1}}
\newcommand{\reffig}[1]{Fig.~\ref{fig:#1}}
\newcommand{\reftab}[1]{Table~\ref{tab:#1}}

\newcommand{\ea}{{\em et al }}

\newcommand{\pose}[2]{^{#1}\mathbf{T}_{#2}}
\newcommand{\img}[1]{\mathbf{I}_{#1}}
\newcommand{\pt}{\mathbf{p}}

\DeclareMathOperator*{\argmax}{arg\,max}

\begin{document}

\maketitle
\thispagestyle{empty}
\pagestyle{empty}


\begin{abstract}

In this paper, a self-supervised model that simultaneously predicts {\em a sequence of future frames} from video-input with a novel spatial-temporal attention (ST) network is proposed. The ST transformer network allows constraining both temporal consistency across future frames whilst constraining consistency across spatial objects in the image at different scales. This was not the case in prior works for depth prediction, which focused on predicting a single frame as output. The proposed model leverages prior scene knowledge such as object shape and texture similar to single-image depth inference methods, whilst also constraining the motion and geometry from a sequence of input images. Apart from the transformer architecture, one of the main contributions with respect to prior works lies in the objective function that enforces spatio-temporal consistency across a sequence of output frames rather than a single output frame. As will be shown, this results in more accurate and robust depth sequence forecasting. The model achieves highly accurate depth forecasting results that outperform existing baselines on the KITTI benchmark. Extensive ablation studies were performed to assess the effectiveness of the proposed techniques. One remarkable result of the proposed model is that it is implicitly capable of forecasting the motion of objects in the scene, rather than requiring complex models involving multi-object detection, segmentation and tracking.

\end{abstract}

\section{Introduction}
\label{sec:new-intro}

Decision-making is a critical process for creating successful autopilots for autonomous driving. This requires a thorough analysis of both the present and future state of the scene. Predicting what might happen in the future will help to make a better-informed decisions. In this paper, a model that outputs the depth of the scene is proposed for both the present (depth inference) and future (depth forecasting) state of the scene, that is obtained through a self-supervised learning approach. Such an approach should enable higher level decision algorithms to make more informed choices based on accurate depth information and predictions.
\\

State-of-the-art approaches, such as~\cite{Watson2021,guizilini2022multi,boulahbal2022forecasting}, have developed models that output a single depth image. The underlying model is then used to perform inference or forecasting tasks separately. These approaches are, however, limited because they cannot enforce spatio-temporal consistency in the output, as they do not predict a sequence. By introducing a model that predicts a sequence of depth images, the model proposed here can apply motion and geometric constraints to the output which improves the accuracy and sharpness of the forecasting and forces the predicted images to be more deterministic (ie. it does not average across possible future outcomes as in prior works). 

\begin{figure}
    \centering
    \includegraphics[width=0.5\textwidth]{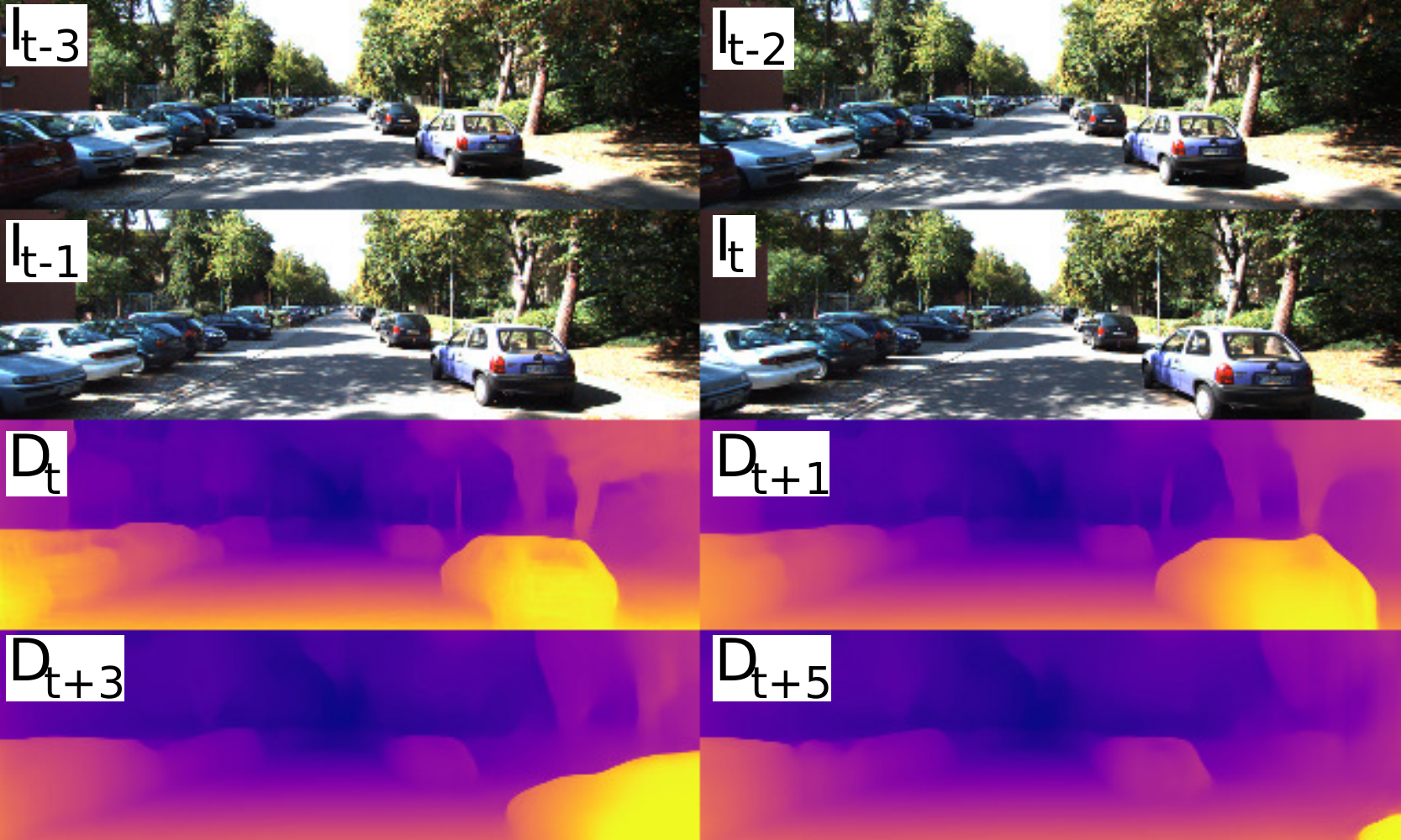}
    \caption{Qualitative performance of the proposed method. The spatially-temporally consistent depth sequence output by the network is shown, including both the current and future depth of the scene. The network uses $4$ context frames and is able to produce an accurate depth sequence.}
    \label{fig:my_label}
\end{figure}

The majority of self-supervised monocular depth inference methods~\cite{Eigen,boulahbal2022instance,safadoust2021self,lee2021attentive,Wang2021,Ranjan2019,Gordon2019,Godard2017,Zhou2017,Godard2019,Johnston2020,Rares2020} rely on a single frame as input. While this approach is effective at leveraging prior knowledge such as object shape and textures, it is limited in its ability to learn the geometry and the motion of the scene. By contrast, using multiple frames~\cite{Watson2021,guizilini2022multi,9864127} as input has the potential to provide a more comprehensive view of the scene and to help the model better understand the relationships between objects and their motions.

Depth forecasting self-supervised methods~\cite{Mahjourian2017,Qi2019,Hu2020,boulahbal2022forecasting}, on the other hand, often produce a blurry depth map that represents the mean of all possible future scenarios~\cite{boulahbal2022forecasting}. This approach fails to produce an accurate depth, which limits its usefulness in decision-making contexts.

To address these limitations, a self-supervised model is proposed that can simultaneously output a depth sequence encompassing inference and forecasting. By using multiple image frames as input and output, the model can learn about the geometric consistency of the scene, which enables it to predict more accurate depth sequences as output. The proposed model enforces a spatio-temporal consistency in the output depth sequence by warping neighboring images onto the target image using a geometric and photometric warping operator that depends on the output depths. As will be detailed further in \refsec{objectivefn}. The results show that this effectively constraints the output depth forecasting to choose the most probable outcome of the future depth, instead of using the mean of all outcomes, avoiding the issue of blurry depth maps and leading to more precise depth. 

As a result, the proposed approach produces an accurate depth sequence. In summary, the contributions of the proposed method are:

\begin{figure*}
    \centering
    \includegraphics[width=\textwidth]{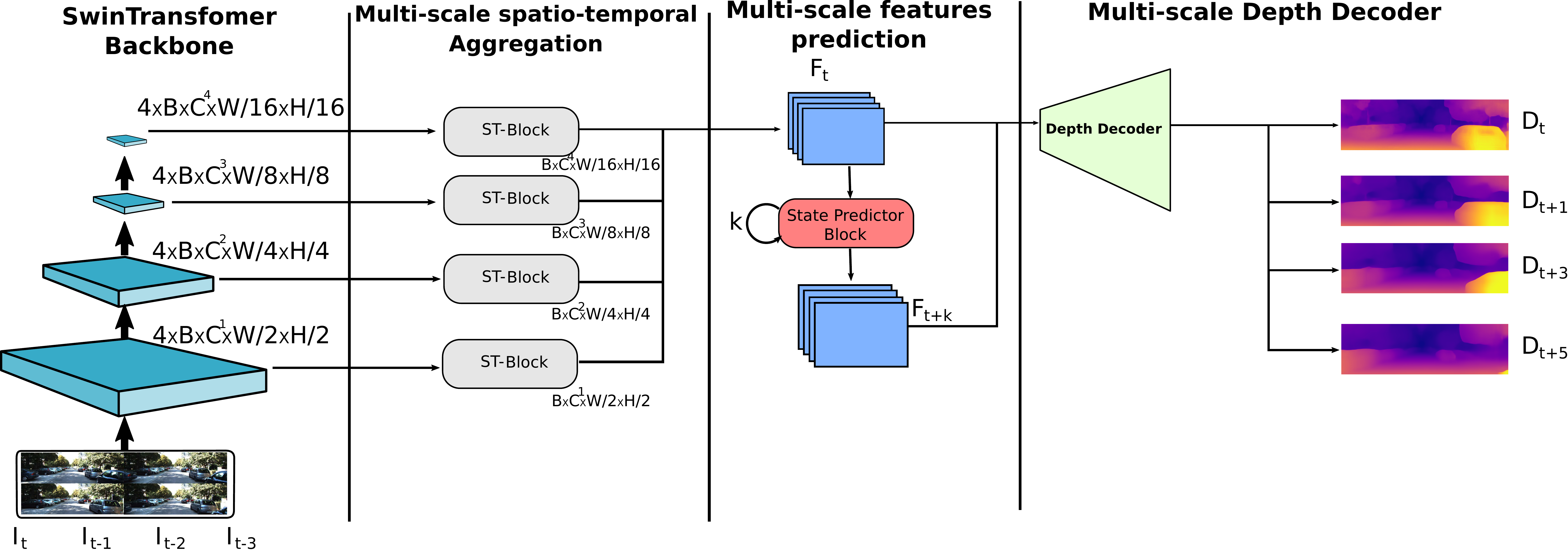}
    \caption{Architecture of the proposed method. The network comprises four stages. Firstly, the spatial feature of each frame is extracted using a SwinTransformer backbone shared across the context frames. Secondly, the features are correlated spatio-temporally using the ST-block shown in \reffig{spblock}. Thirdly, a learned function $f$ is used to transition from $F_{t+k-1}$ to $F_{t+k}$, and this module consists of SwinTransformer blocks as well. Finally, the depth decoder employs skip connections to utilize multi-scale features and outputs 4 depth states: $(\mbf{D}_{t}, \mbf{D}_{t+1} ,\mbf{D}_{t+3} \mbf{D}_{t+5})$}
    \label{fig:architecture-chapter3}
\end{figure*}

\begin{itemize}
    \item A self-supervised model that predicts a spatially and temporally consistent depth sequence that captures both present and future depth information, allowing for more comprehensive and accurate depth.
    \item A transformer-based multi-frame architecture that implicitly learns the geometry of the scene in an image-based end-to-end manner. Interestingly, the proposed model is capable of forecasting the motion of objects in the scene, even in the absence of explicit motion supervision.
    \item The method achieves highly accurate depth forecasting results that outperform existing baselines in the KITTI~\cite{Geiger2012CVPR} benchmark.
    \item Improved generalization for depth inference tasks over SOTA.     
    \item A comprehensive analysis of the proposed method is conducted through several ablation studies.

\end{itemize}

\section{Related work}

To clarify ambiguous terminology found across different papers in the related work, the term “prediction” will be considered here to encompass both depth inference and future forecasting. The term "inference" will be reserved for the prediction at time $t$ and "forecasting" will be used for future predictions $(t+1:t+k)$.

\subsection{Depth inference}
Self-supervised learning from videos has been successful for depth inference. Pioneering work by Zhou \ea~\cite{Zhou2017} introduced the idea of optimizing the pose and depth network jointly using image warping and photometric loss. However, this task is inherently ill-posed (recovering 3D from a 2D image) and several subsequent works have addressed challenges to improve the performance of self-supervised depth inference~\cite{Zhou2017,Gordon2019,Shu2020,Bian2019,Chen2019,Rares2020,Wang2021,Watson2021,chen2019self,Ranjan2019,Klingner2020,Vijayanarasimhan,Lee2019,Johnston2020,johnston2020self}. 

Most of these models do not leverage any temporal information during evaluation, limiting their performance. The use of multiple frames has the potential to improve the performance of depth networks, as determining the geometry of the scene would be possible. Manydepth~\cite{Watson2021}, MonoRec~\cite{wimbauer2021monorec} and DepthFormer~\cite{guizilini2022multi} have leveraged cost volumes to enable geometric-based reasoning during inference. However, they rely on the plane sweep process to perform the matching. Instead, the proposed method learns the geometry implicitly using the transformer's attention framework.

\subsection{Depth forecasting}
Several previous works have investigated depth forecasting from monocular videos. Mahjourian \ea~\cite{Mahjourian2017} introduced using forecast depth to generate the next RGB image frame, supervised by ground-truth LiDAR scans. Similarly, Qi \ea~\cite{Qi2019} and Hu \ea~\cite{Hu2020} used multi-modal inputs to forecast future modalities including depth, semantic, and optical flow. However, these methods require ground-truth labels for supervision during both training and testing, limiting their real-world applicability. \cite{boulahbal2022forecasting} proposed a self-supervised transformer-based model for depth forecasting, but it outputs only the mean of possible future depths and cannot accurately forecast depth. In contrast, the proposed method performs depth inference and forecasting simultaneously and outputs only the most probable future depth, leading to significant performance improvement over prior work.

\section{Method}
\subsection{Problem formulation}
The aim of monocular depth inference and forecasting is to predict an accurate depth sequence through the mapping, $\mbf{D}_{t:t+n} =~f(\img{t-k:t}; \bm{\theta})$ where $\img{t-k:t}$ are the $k$ context images and $\mbf{D}_{t:t+s}$ are the $s$ depth target states. In self-supervised learning, this model is trained via novel view synthesis by warping a set of source frames $\mbf{I}_{src}$ to the target frame $\mbf{I}_{tgt}$ using the learned depth $\mbf{D}_{tgt}$ and the target to source pose $\pose{src}{tgt} \in \mathbb{SE}[3]$~\cite{jaderberg2015spatial}. The differentiable warping function~\cite{Zhou2017} is defined in as follows:
\begin{equation}
        \widehat{\pt}_{src} \sim \mbf{K} \pose{src}{tgt} \mbf{D}_{tgt}  \mbf{K}^{-1} \pt_{tgt}
\end{equation}
The depth network takes $k=4$ context images as input and produces $tgt=\{0,1,3,5\}$ depth outputs. As the pose network is only used for supervision during training, providing the future images will help the pose network to learn better. Therefore, for each depth state $tgt$, the pose network input is the triplet of images $(tgt-1, tgt, tgt+1)$. It outputs two poses, $\pose{tgt-1}{tgt}$ and $\pose{tgt}{tgt+1}$.

\subsection{Architecture}

The proposed model is related to classic \textit{structure-from-motion}. During training, self-supervision is achieved by using an image warping function, and two networks are used: a pose network and a depth network. At test time, only the depth network is used and the pose network is discarded.
\subsubsection{The depth network}
\reffig{architecture-chapter3} shows the architecture of the proposed method. The depth network uses $k=4$ context inputs. The architecture comprises four stages: 

\textbf{1. Spatial feature extraction:} SwinTransformer backbone~\cite{liu2021swin} is used to extract the features of each frame. The swin-tiny variant is used with a number of layers : [$2, 2, 6, 2$], with depths of [$3, 6, 12, 24$], a patch embedding channel of $7$, and an embedding dimension of $96$. It is pretrained on the ImageNet dataset~\cite{deng2009imagenet}. See~\cite{liu2021swin} for more details. This feature extractor is shared across the context frames. The feature map at each scale is extracted as input for the next module. As the purpose of this module is to extract spatial information only, calculating the gradient for only one context frame is sufficient. Experimentally, no differences were observed between calculating the gradient for all four frames and only one frame. Therefore, backpropagation is carried out only on the first frame to minimize the memory footprint.
    
\begin{figure}
    \centering
    \includegraphics[width=0.5\textwidth]{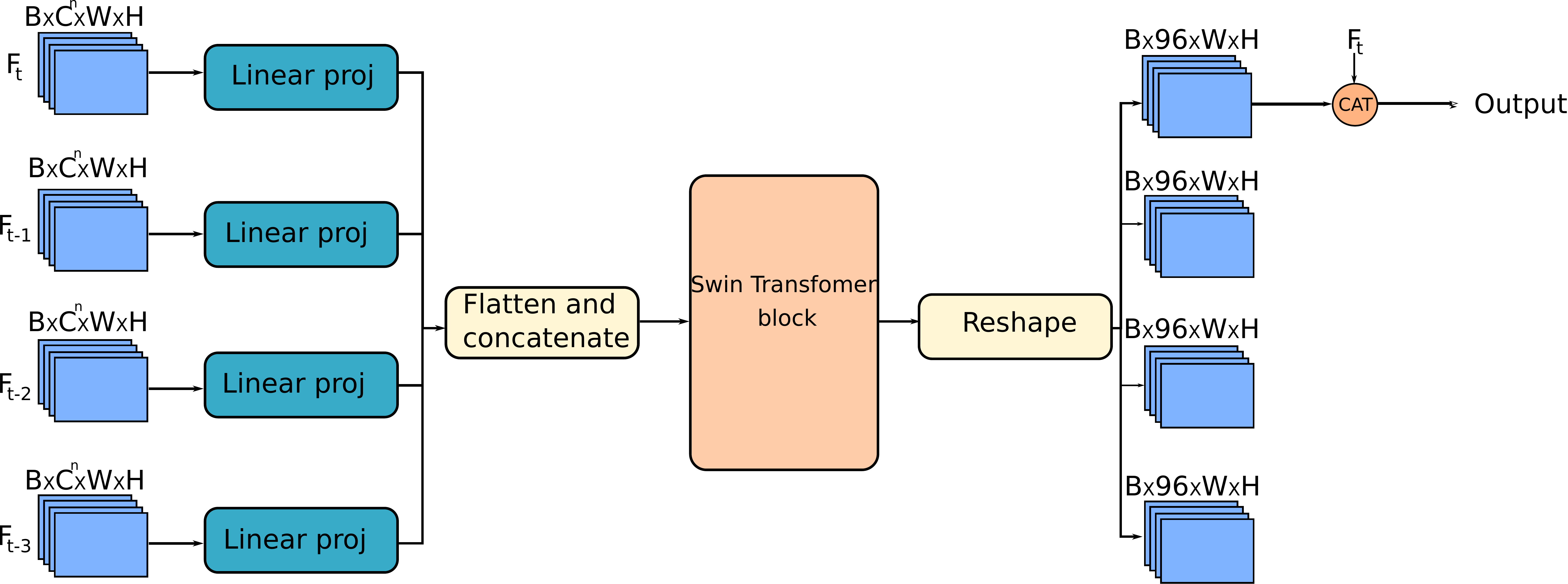}
    \caption{Architecture of a multi-scale spatio-temporal aggregation network using linear projection and SwinTransformer layers for feature spatio-temporal correlation.}
    \label{fig:spblock}
\end{figure}

\textbf{2. Multi-scale spatio-temporal aggregation:} Next, the features are correlated spatio-temporally using the proposed novel ST-block. \reffig{spblock} shows the architecture of this fusion block. At each feature scale, each feature map of each frame is projected using a \textit{Conv2D} with $kernal=1$ outputting an embedding of dimension $96$. These features are concatenated as a sequence of patches to construct embeddings that will be used as input to the transformers. This sequence is then provided to the transformer~\cite{liu2021swin}. This block has a depth of $2$ and embeddings of $96$ and the number of heads at each scale is [$3, 6, 12, 24$] from high to low resolution. The attention map performs the spatio-temporal correlation of these features. The sequence is reshaped to its original shape and the first feature map is contacted with the context feature $F_t$. Finally, another projection layer outputs the spatio-temporal features to recover the channel to $C^n$.

\begin{figure*}
    \centering
    \includegraphics[width=\textwidth]{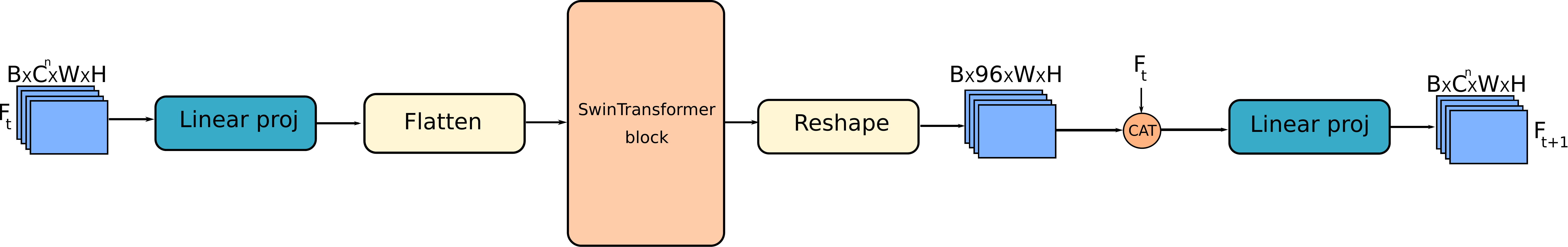}
    \caption{SwinTransformer-based state predictor block. The input feature map $F_t$ is projected onto an embedding dimension of size $96$ and flattened into patches for the SwinTransformer block. The output is reshaped and concatenated with the input using a skip connection. A linear projection generates the features of $Ft+1$ with size $B \times C_n \times W \times H$ where $C_n$ is the original channel}
    \label{fig:temporal-block}
\end{figure*}

\textbf{3. Multi-scale feature prediction:} A transition function $f$ is used to relate each feature to a state in the output sequence. At each scale, this learned function $f$ is used to transition from $F_{t+k-1}$ to $F_{t+k}$. This function is recursive and defined as:
        \begin{align}
            F_{t+k} &= f(F_{t+k-1}) 
        \end{align}
This module is composed of SwinTransformer blocks~\cite{liu2021swin}. It is shared and used recursively across all $n$ frames to be forecast. \reffig{temporal-block} shows the architecture of the state predictor block. The input feature map $F_t$ is projected to have an embedding dimension of $96$. This map is flattened to patches of size $1$ to be used as input to the SwinTransfomer block. Similarly, this block has a depth of $2$, embeddings of $96$ and the number of heads of each scale is [$3, 6, 12, 24$] from high to low resolution. The output is reshaped to its original dimensions and concatenated with the input with a skip connection. A linear projection is used to obtain the features of $F_{t+1}$ with size: $B\times C^n \times W \times H$. 

\textbf{4. Depth decoder: } This module is shared across all state features. It consists of Transposed2DConvolution with ReLU as activation and a kernel size of $k=3$, which is similar to~\cite{Godard2019}. Skip connections are employed since the previous stage outputs multi-scale features. In this method, four depth states are output: $(\mbf{D}_{t}, \mbf{D}_{t+1} ,\mbf{D}_{t+3}, \mbf{D}_{t+5})$.

\subsubsection{The pose network}
This network is an off-the-shelf model taken from~\cite{Godard2019} that takes the triplet $(tgt-1, tgt, tgt+1)$ and outputs two poses: $\pose{tgt-1}{tgt}$ and $\pose{tgt}{tgt+1}$. This model is used only for self-supervised training and is discarded at evaluation. 

It is worth noting that current state-of-the-art methods utilize a plane sweep approach, such as the one proposed by \cite{Watson2021,guizilini2022multi}, that involve explicitly providing the pose and camera parameters to the depth network and constructing a matching volume during training and evaluation. Alternatively, the proposed method adopts a different approach that learns this information implicitly. This presents several benefits, most notably the ability for the two networks, the depth and the pose, to operate independently. This independence from the pose network and camera parameters is particularly significant, as it allows the proposed network to generalize better and perform more robustly. Empirical evidence supporting this claim is presented in~\refsec{generalization}, where the experimental results demonstrate the superiority of the proposed approach.

\subsection{Objective functions}
\label{sec:objectivefn}
As the self-supervision is done by reconstructing the frames $\mbf{I}_{tgt}$ such as $tgt \in \{0,1,3,5\}$ using the depth and the pose with the warping, this can be formulated as a maximum likelihood problem: 
 \begin{align}
 \footnotesize
     \widehat{\bm{\theta}} &= \argmax_{\bm\theta \in \bm{\Theta}} L(\mbf{I}_{t+5}, \mbf{I}_{t+3} ,\mbf{I}_{t+1} ,\mbf{I}_{t}| \mbf{I}_{t-4:t+6}; \bm{\theta} ) \\ \notag &\equiv	 \argmax_{\bm\theta \in \bm{\Theta}} \sum_{m} P_{model}(\mbf{I}_{t+n}^m| \mbf{I}_{c}^m,\mbf{I}_{f}^m)
 \end{align} 
 $\bm\theta$ is the model parameters, $\mbf{I}_{c}^m$ are the context frames that will be provided to the depth network and $\mbf{I}_{f}^m$ are the future frames that will be provided to the pose network. Similar to~\cite{Godard2019}, the photometric loss and structural similarity index measure SSIM~\cite{wang2004image}, along with the depth smoothness are used to optimize the parameters. 
\begin{align}
\label{eq:ssim}
\footnotesize
  pe(\img{tgt}, \widehat{\mbf{I}}_{(tgt\pm1 \rightarrow tgt)} )= \sum_{\pt} \big[&(1 - \alpha) \textup{ SSIM}[ \img{tgt}(\pt) - \widehat{\mbf{I}}_{(tgt\pm1 \rightarrow tgt)}(\pt)]  \\ \notag
      & +  \alpha | \img{tgt}(\pt) - \widehat{\mbf{I}}_{(tgt\pm1 \rightarrow tgt)}(\pt) |\big]
\end{align}
Such that $\widehat{\mbf{I}}_{(tgt\pm1 \rightarrow tgt)}$ is reconstructed from two views : $\img{tgt-1}$ and $\img{tgt+1}$. Similar to~\cite{Godard2019}, the minimum projection loss of the two frames is used to handle occlusions leading to: 
\begin{align}
\mathcal{L}_\textup{ph}(\img{tgt})= min\big[pe(\img{tgt}, \widehat{\mbf{I}}_{(tgt-1 \rightarrow tgt)} ) , pe(\img{tgt}, \widehat{\mbf{I}}_{(tgt+1 \rightarrow tgt)} ) \big]
\end{align}
To further improve the training, outlier rejection is performed. Similar to~\cite{Godard2019} this is done using auto-masking which is defined as: 

\begin{align}
\footnotesize
    \mu  = \big[ min_{tgt}\big(pe(\img{tgt}, \widehat{\mbf{I}}_{(tgt\pm 1\rightarrow tgt)}), pe(\img{tgt}, \mbf{I}_{(tgt\pm1)}) \big) \big]
\end{align}
where [ ] is the Iverson bracket. $\mu$ is set to only include the loss of pixels where the re-projection error
of the warped image $\widehat{\mbf{I}}_{(tgt\pm 1\rightarrow tgt)}$ is lower than that of the original, unwarped image $\mbf{I}_{(tgt\pm1)}$. An edge-aware gradient smoothness constraint is used to regularize the photometric loss. The disparity map is constrained to be locally smooth.  
\begin{align}
\footnotesize
    \begin{aligned}
      \mathcal{L}_{s}(D_{tgt}) = \sum_{p} \big[ &| \partial_x D_{tgt}(\pt) | e^{-|\partial_x \mbf{I}_{tgt}(\pt)|}    +
      &| \partial_y D_{tgt}(\pt) | e^{-|\partial_y \mbf{I}_{tgt}(\pt)|} \big] 
    \end{aligned}\label{eq:loss-disp-smoothness}
\end{align}
Temporal consistency is enforced during training through a loss function that enforces geometric constraints simultaneously across multiple output frames, namely $\mbf{D}_{t+5}$, $\mbf{D}_{t+3}$, $\mbf{D}_{t+1}$, and $\mbf{D}_{t}$ via a warping function. The image-based loss function minimizes the pair-wise photometric consistency by warping neighboring images for each central target ($\img{t+5}$, $\img{t+3}$, $\img{t+1}$, and $\img{t}$). The warping function depends on the output depth and pose outputs from the network. This constrains the model to respect image consistency between these frames. For example, $\img{t+1}$ is minimized with respect to the warped $\widehat{\mbf{I}}_{t}$ and the warped $\widehat{\mbf{I}}_{t+2}$. The gradient locality problem~\cite{Zhou2017} is handled using a pyramid of depth outputs, and the optimization is done on all these levels. The final loss function is defined as: 
\begin{align}
      \mathcal{L} = \frac{1}{m}\sum_{m}\sum_{tgt}\sum_{l=1}^{l=4} \mu \mathcal{L}_\textup{ph}(\img{tgt}) + \alpha_s \mathcal{L}_{s}(D_{tgt})
\end{align}
where $m$ is the batch size, $tgt\in{0,1,3,5}$ and $l$ represents the multiscale output depth.
 \section{Results}

\begin{figure*}
    \centering
    \includegraphics[width=\textwidth]{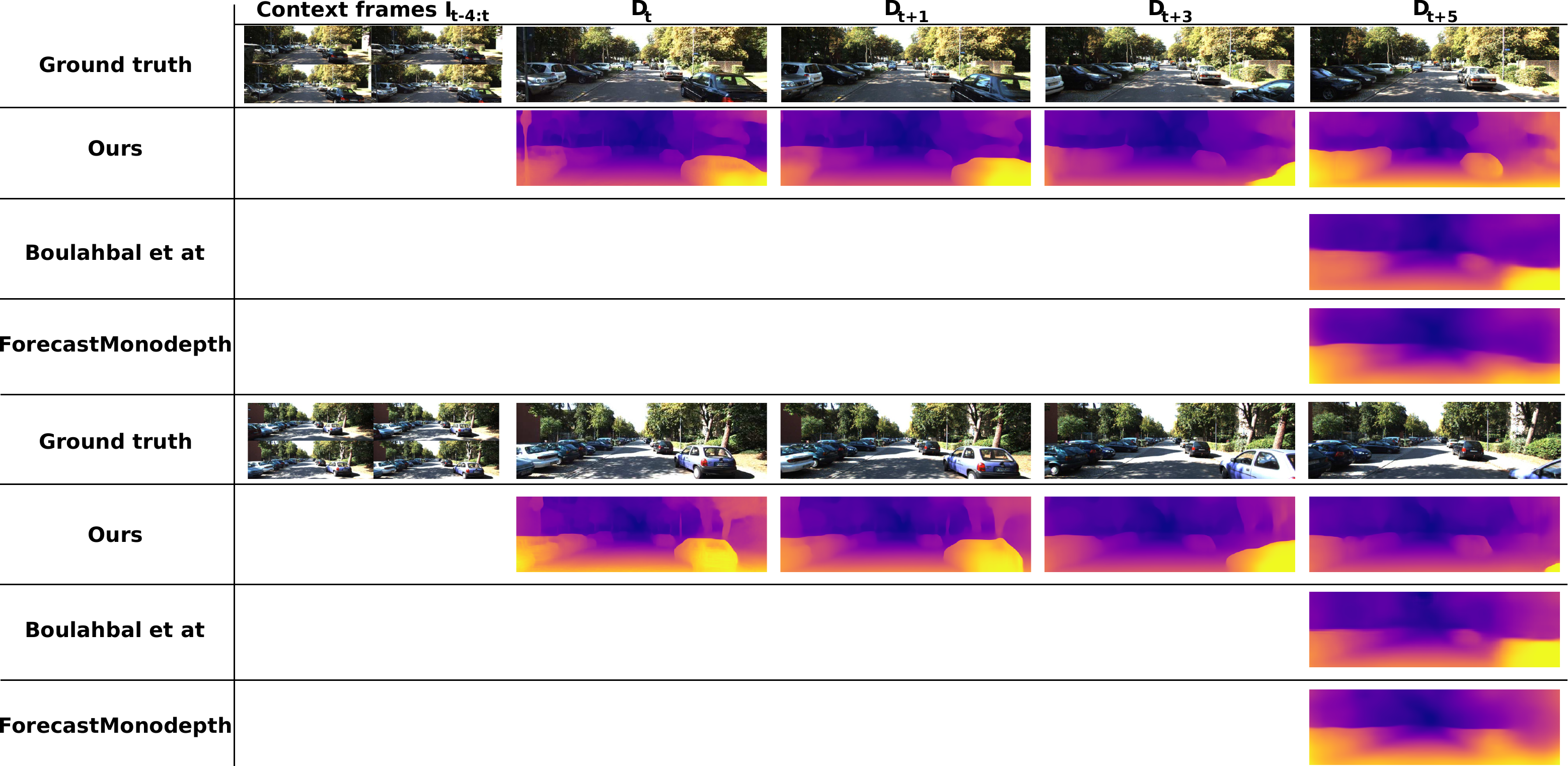}
    \caption{ Qualitative comparison of the proposed method and the prior work on KITTI Eigen test benchmark. The proposed method is able to generate an accurate future depth sequence that exhibits significantly more details compared to the prior work. The depth map generated by the proposed method is remarkably sharp and not blurry. This superior performance can be attributed to the fact that the proposed method was specifically trained for depth inference with spatio-temporal consistence across the forecast range, resulting in an enforced deterministic output. As a result, the proposed approach predicts the most probable future instead of averaging all possible futures, as done in the prior work.}
    \label{fig:qualitative_perf}
\end{figure*}

\label{sec:new-results}
\subsection{Experimental setup}

\textbf{Datasets:} The KITTI benchmark~\cite{Geiger2012CVPR} is the defacto benchmark for evaluating depth methods. The Eigen \ea~\cite{Eigen} is used with Zhou \ea\cite{Zhou2017} preprocessing to remove static frames. In order to test the generalization of the method, the Cityscapes~\cite{cordts2016cityscapes} and the Robotcar~\cite{RobotCarDatasetIJRR} datasets are used. The Cityscapes dataset does not provide ground truth LiDAR depth and uses the classical SGM method~\cite{4359315} to obtain the depth. The LiDAR depth is projected onto the image to obtain the ground truth for the Robotcar dataset.

\textbf{Baselines:} Several depth inference method were used for the comparison~\cite{Watson2021,guizilini2022multi}. For forecasting, a comparison is made only with methods that perform self-supervision with respect to frame $5$ as done in~\cite{boulahbal2022forecasting}. To test the performance of the method with respect to dynamic objects, the analysis provided in~\cite{boulahbal2022instance} is used.

\textbf{Hyperparameters:} The networks are trained for $6$ epochs, with a batch size of $4$. The Adam optimizer~\cite{kingma2014adam} is used with a learning rate of $lr = 10^{-4}$ and $(\beta_1, \beta_2) = (0.9, 0.999)$. The SSIM weight is set to $\alpha = 0.15$ and the smoothing regularization weight to $\alpha_s = 0.001$. $l = 4$ scales are used for each output of the decoder. At each scale, the depth is upscaled to the target image size. The input images are resized to $192 \times 640$. Two data augmentations were performed: horizontal flips with probability $p = 0.5$ and color jitter with p = 1. The activation of depth decoder, $\sigma$, is re-scaled to obtain the depth $D = \frac{1}{a \sigma +b}$, where $a$ and $b$ are chosen to constrain $D$ between $0.1$ and $100$ units for training and $(0.5,100)$ for evaluation, similar to~\cite{boulahbal2022instance}. The scale ambiguity is resolved using median scaling, similar to the prior work~\cite{Godard2019,Watson2021}.

\begin{table*}
\centering
\footnotesize
\begin{tabular}{|l|c||c c c c| c c c|}
\hline 
Predicted frame & Method &Abs Rel& Sq Rel&  RMSE & RMSE log  &
$\delta<1.25$ &
$\delta<1.25^2$ &
$\delta<1.25^3$\\
\hline \hline
$\mbf{t=0}$ 
& SfMLearner~\cite{Zhou2017} &0.198 & 1.836& 0.275  & 6.565 & 0.718 & 0.901 & 0.960\\
& Yang \ea~\cite{yang2018unsupervised} & 0.182 & 1.481  & 0.267& 6.501 & 0.725 & 0.906 & 0.963\\
& GeoNet~\cite{Yin2018} & 0.155 & 1.296 & 5.857  & 0.233 & 0.793 & 0.931 & 0.973\\
& CC~\cite{Ranjan2019} & 0.140 & 1.070  & 5.326  & 0.217 &0.826 &0.941 &0.975 \\

& Monodepth2~\cite{Godard2019}& 0.115 & 0.903 & 4.863 & 0.193 & 0.877 & 0.959 & 0.981\\
& Lee \ea~\cite{lee2021attentive}  &  0.113 & 0.835 & 4.693 & 0.191 & 0.879 & 0.961 & 0.981 \\
& PackNetSfm~\cite{Rares2020} & 0.111 & 0.785 & 4.601 & 0.189 & 0.878 & 0.960 & 0.982\\
& Manydepth~\cite{Watson2021}& \textbf{0.098} & \textbf{0.770} & \textbf{4.459} & \textbf{0.176} & \textbf{0.900 }& \textbf{0.965} & \textbf{0.983} \\

& \textbf{Ours} & 0.110  &   0.805  &   4.678  &   0.187  &   0.879  &   0.961  &   0.983 \\
\hline
$\mbf{t=1}/ 0.1sec$ & \textbf{Ours} & 0.121  &   0.989  &   5.026  &   0.203  &   0.863  &   0.951  &   0.978 \\
\hline
$\mbf{t=3}/ 0.3sec$ & \textbf{Ours} & 0.146  &   1.295  &   5.493  &   0.227  &   0.824  &   0.935  &   0.971 \\
\hline
$\mbf{t=5}/ 0.5sec$ 
&
ForecastMonodepth2~\cite{boulahbal2022forecasting} & 0.201  &   1.588   &   6.166   &   0.275 &   0.702 &  0.897  &   0.960\\

& Boulahbal~\ea~\cite{boulahbal2022forecasting}&  0.178  &  1.645   &   6.196 &   0.257  &   0.761  &   0.914  &   0.964  \\

& Ours & \textbf{0.165}  &   \textbf{1.489}  &   \textbf{5.805  }&  \textbf{ 0.245}  &   \textbf{0.792}  &   \textbf{0.921 } &  \textbf{ 0.964 }\\
\hline 
\end{tabular}
\caption{Quantitative performance of the proposed method on the KITTI~\cite{Geiger2012CVPR} eigen~\cite{Eigen} benchmark for the frames $D_t, D_{t+1}, D_{t+3}, D_{t+5}$. for Abs Rel, Sq Rel, RMSE and RMSE log lower is better. For $\delta<1.25$, $\delta<1.25^2$, $\delta<1.25^3$ higher is better. The proposed method is able to output an accurate depth at different time steps. The performance of the future depth is even comparable to depth inference method that have access to the target frame.  }
\label{tab:results-1}
\end{table*}




\begin{table}
\begin{center}
\footnotesize
\resizebox{\columnwidth}{!}
{
\begin{tabular}{|c|c|c|c|c|c|}
    \hline
    Evaluation& Model & Abs Rel & Sq Rel & RMSE & RMSE log  \\ 
    \hline
    All points mean &
    ManyDepth \cite{Watson2021} & 0.098 & 0.770 & 4.459 & 0.176   \\
    &
    Boulahbal~\ea~\cite{boulahbal2022instance} &  0.110  &   0.719  &   4.486  &   0.184 \\
 &
    Ours & 0.110  &   0.805  &   4.677  &   0.187  \\
 &

    Ours + stereo &  0.107  &   0.751  &   4.805  &   0.189 \\

    \hline
    \hline 
    Only dynamic &
    ManyDepth \cite{Watson2021} &   0.192  &   2.609  &   7.461  &   0.288   \\
    &
    Boulahbal~\ea~\cite{boulahbal2022instance} & 0.167  &   1.911  &   6.724 &   0.271 \\
&
    Ours &  0.178  &   2.089  &   6.963  &   0.278  \\
     &

    Ours + stereo &  0.155  &   1.668  &   6.401  &   0.260  \\
    
    \hline
    \hline 
    Only static &
    ManyDepth\cite{Watson2021} &  0.085  & 0.613  &   4.128 &   0.150    \\ &
    
    Boulahbal~\ea~\cite{boulahbal2022instance} & 0.101  & 0.624  &   4.269  &   0.163 \\
    &
    Ours  & 0.099  &   0.684  &   4.462  &   0.165\\
     &

    Ours + stereo & 0.099  &   0.684  &   4.679  &   0.173 \\
    \hline
    \hline
    Per category mean &
    ManyDepth\cite{Watson2021} &   0.139   &  1.611 & 5.794   &  0.219\\ &
    Boulahbal~\ea~\cite{boulahbal2022instance} & 0.134  &  1,267  &   5,496 & 0,217  \\
     &
    Ours & 0.138  &   1.386  &  5.712  &   0.222 \\
     &

    Ours + stereo & 0.127 &  1.176  &  5.540 &   0.217 \\
    \hline
    \end{tabular}
}
\caption{Quantitative performance comparison for dynamic and static objects at $t=0$0. The proposed method outperforms the SOTA~\cite{Watson2021} on the dynamic objects. The stereo variant is the best model for the dynamic and the per category mean.}
\label{tab:staticVSdynamic}

\end{center}
\end{table}
\subsection{Multi-step depth forecasting results}
The findings of the study reveal that the proposed method exhibits a faster convergence rate, with a reduced number of epochs compared to prior works. Specifically, the proposed method achieves convergence in just $6$ epochs, whereas previous approaches required $20$ epochs.

It is of particular interest to examine \reftab{results-1}, which displays the performance of the proposed method across different predicted future steps. Notably, it can be observed that as time progresses, the uncertainty of the future increases, leading to larger errors in the predictions. 

Furthermore, \reftab{results-1} presents the results of comparing the depth forecasting of the proposed method with prior works. As expected, the proposed method outperforms the prior work, with a significant gap of $\Delta AbsRel = 7.3\%$. This finding is further substantiated by the qualitative observations portrayed in \reffig{qualitative_perf}, where the generated output depth maps produced by our proposed method demonstrate superior precision and intricate details when compared to prior approaches. The forecasted depth at the different time steps is even comparable to other methods that does depth inference and have access to the target frame image. $t=1$ it is better than~\cite{Ranjan2019,Yin2018,yang2018unsupervised,Zhou2017}. and $t=5$ is better than~\cite{yang2018unsupervised,Zhou2017}. This demonstrates that the enforcing the spatio-temporal consistency results in an accurate depth sequence.

\subsection{Handling dynamic objects}
\label{sec:dynamic-obj}

In the interest of conducting a thorough analysis of the proposed method, it is important to consider its ability to handle dynamic objects in the scene. One limitation of a previous approach (\cite{Watson2021}) was its inability to handle such objects, and thus we conducted an experiment to address this issue.

The proposed model will be evaluated against~\cite{Watson2021,boulahbal2022instance} using the methodology introduced in~\cite{boulahbal2022instance}. Furthermore, a variant of the proposed method is employed which leverages stereo images during training. The pose network is completely discarded, and the extrinsic parameters of the stereo pair are used to warp the one view into the other. This variant model operates under the assumption of a rigid scene, thereby avoiding any issues related to warping dynamic objects.


The results, presented in~\reftab{staticVSdynamic}, show that the proposed method outperforms the ManyDepth baseline on dynamic objects with a significant improvement of $\Delta AbsRel=13.0\%$, and it has a comparable result when the unbiased per category mean is used. Although the proposed variant with stereo images during evaluation performs almost equally well for the static scenes, most of the improvement is observed on the dynamic objects. These findings suggest that although the proposed model is not explicitly trained on dynamic objects, it is able to learn their dynamics implicitly. One possible explanation for this is that the as model utilizes multi-scale attention, which allows it to capture the motion of dynamic objects even without being specifically supervised to do so. This highlights the effectiveness of attention mechanisms in capturing spatio-temporal dependencies and modeling the dynamics of the scene. Overall, these experiments highlight the importance of handling the dynamic objects in depth inference and forecasting.

\subsection{Depth inference generalization}
\label{sec:generalization}

In order to compare the proposed architecture with other methods that perform single depth-image inference, it is proposed to train the model only for this (output only depth at $D_t$). The comparison is only made with respect to methods that leverage multi-fame input for the depth network. Prior methods~\cite{Watson2021,guizilini2022multi} perform a plane sweep operation to compute a cost volume. The plane sweep algorithm explicitly uses the pose of the scene and requires the camera intrinsic parameters. The proposed depth network model, on the other hand, performs the matching implicitly using the transformers and does not depend on any other network.

\begin{figure}
    \centering
    \includegraphics[width=0.5\textwidth]{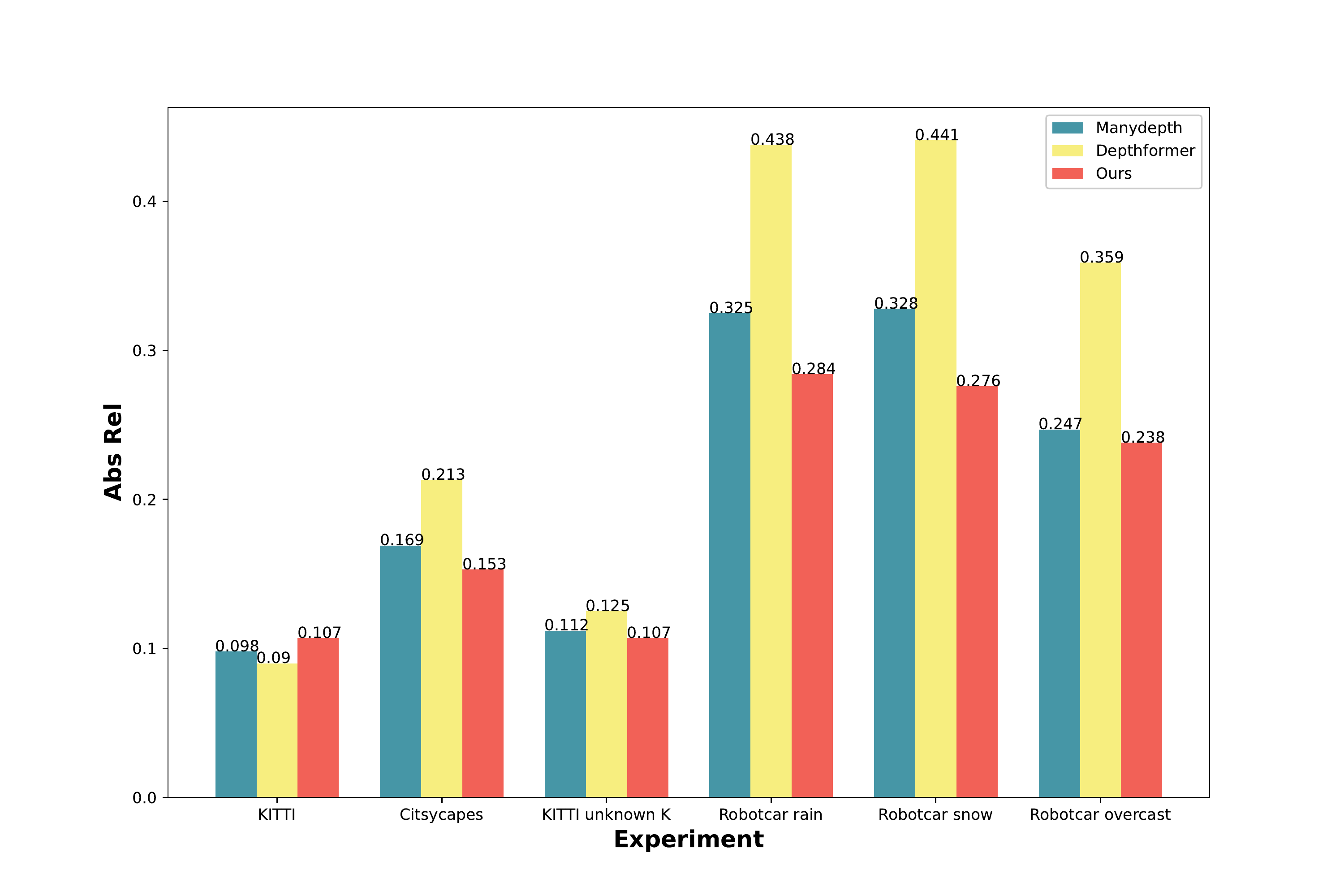}
    \caption{Depth inference generalization study. The proposed architecture is compared to ManyDepth and DeptFormer on different generalization scenarios: Domain gap evaluation on Cityscapes, sensitivity to camera parameters, and weather perturbations on the Robotcar dataset. As shown, the proposed method outperforms the baselines in all three generalization settings, suggesting its ability to generalize well to different scenarios.}
    \label{fig:generalization_study}
\end{figure}

\reftab{results-1} shows the comparison results of the proposed method with ManyDepth~\cite{Watson2021} on KITTI benchmark~\cite{Geiger2012CVPR}. As expected, the models that explicitly use the pose information have better performance on the KITTI benchmark for depth inference. These assessments, however, do not evaluate the ability of the networks to generalize to new scenes. Therefore, a generalization study was performed to better assess the models in this respect:
\begin{itemize}
    \item \textbf{Testing the domain gap:} The models pre-trained on the KITTI dataset are directly evaluated on the Cityscapes~\cite{cordts2016cityscapes} dataset without retraining.  
    \item \textbf{Testing the sensibility to the camera parameters:} The focal length of the camera is replaced with a focal length $f=1$ and the optical center is chosen as $(\frac{W}{2},\frac{H}{2})$. This evaluation is performed on the KITTI dataset. 
    \item \textbf{Testing weather perturbation:} The evaluation is done on $3$ sequences of the Robotcar dataset: overcast, snow and rain sequences.
\end{itemize}
As observed in~\reffig{generalization_study} the proposed method outperforms the baselines for these generalization settings. This suggests that while the baselines are able to perform better on the KITTI dataset, they do generalize better in other settings. This could be explained by the fact that the generalization of these methods depends on both the generalization of the pose and depth network. The proposed model, on the other hand, performs matching implicitly using the attention of transformers and does not depend on any other network, which makes it less sensitive to variations in pose and camera parameters. 

\subsection{Ablation study}

\begin{figure}
    \centering
    \includegraphics[width=0.5\textwidth]{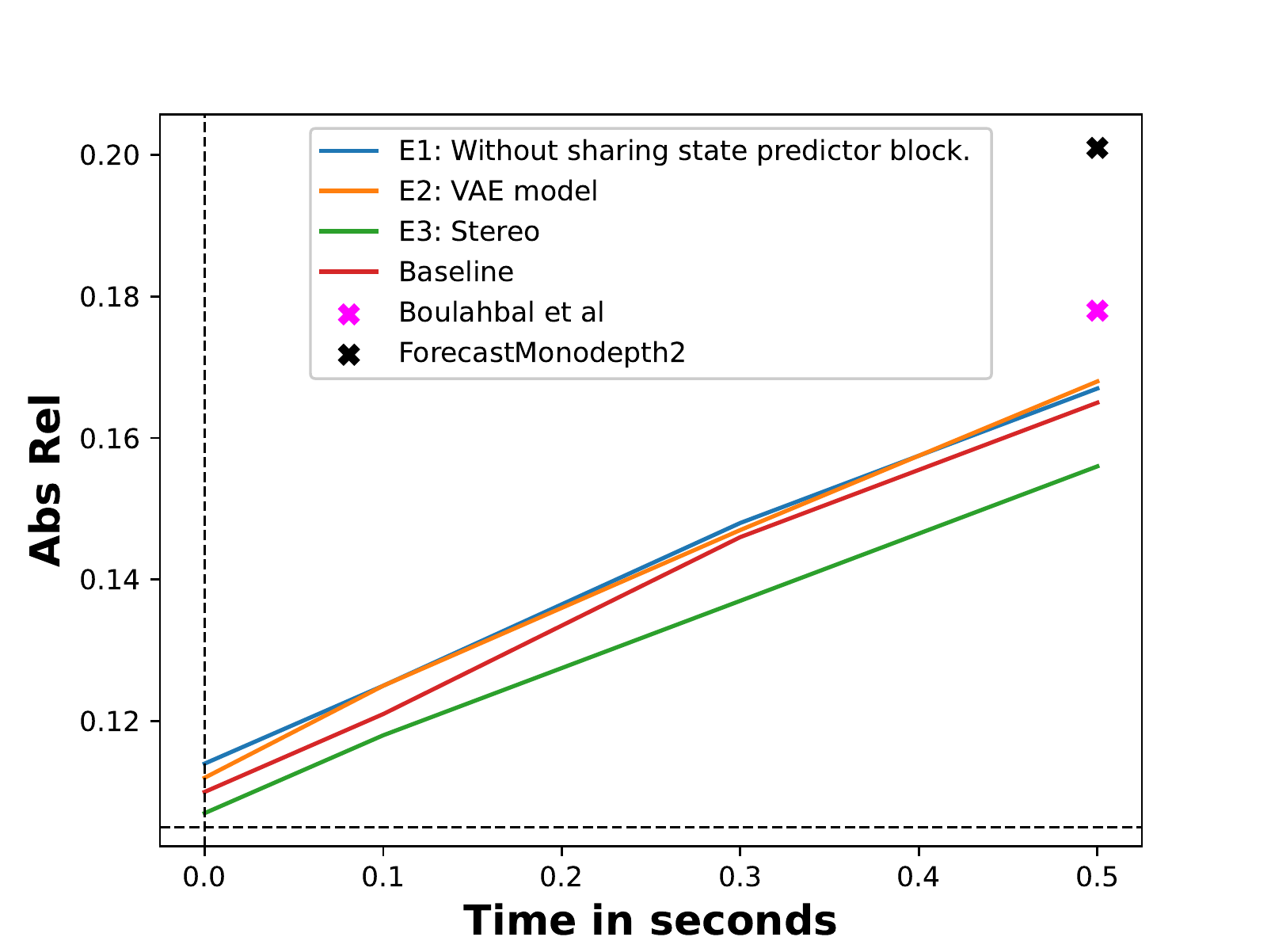}
    \caption{Ablation studies for improving depth forecasting performance. The \textit{Abs~Rel} performance of various evaluated models is shown in the figure. (i) E1, tests the model without sharing the state predictor block. (ii) E2, involves the use of a VAE model to output multi-hypothesis future depth (iii) E3, assess the model with the stereo pose.}
    \label{fig:ablations-chap6}
\end{figure}

Several ablations were performed in an effort to improve the performance of the proposed model. Figure~\ref{fig:ablations-chap6} displays the \textit{Abs~Rel} performance of the various evaluated models. Specifically, the following ablation studies were carried out:
\begin{itemize}
    \item E1: Tests the model without sharing the state predictor block. As observed, sharing the state predictor helps the model to output a better depth as multiple passes helps the network to generalize better.
    
    \item E2: This experiment involved the use of a VAE model, where the latent variables of the state predictor block were assumed to follow a Gaussian distribution. The aim of this experiment was to output a multi-hypothesis future depth. However, the first observation we made was that the model collapsed to a single modality and failed to output multiple hypotheses. As the decoder was perturbed with the Gaussian distribution, the output is less accurate to the baseline.

    \item E3: Aim to assess the model with dynamic objects.More details are provided in~\refsec{dynamic-obj}

\end{itemize}
These experiments demonstrate that the proposed method holds several advantages: providing a spatial-temporal consistent depth sequence that represents present and future depth, superior depth forecasting compared to the prior work, and better generalization for depth inference.   

\label{sec:res-gen}

\section{Conclusion}
\label{sec:new-conclusion}
 In conclusion, this paper presents a novel self-supervised model that predicts a sequence of future frames from video-input using a spatial-temporal attention (ST) network. The proposed model outperforms existing baselines on the KITTI benchmark for depth forecasting and achieves highly accurate and robust depth inference results. The novelty of the proposed model lies in its use of a transformer-based multi-frame architecture that implicitly learns the geometry and motion of the scene, while also leveraging prior scene knowledge such as object shape and texture. Furthermore, the proposed model enforces spatio-temporal consistency across a sequence of output frames rather than a single output frame, resulting in more accurate and robust depth sequence forecasting. Several ablation studies were conducted to assess the effectiveness of the proposed techniques. The proposed model provides a significant contribution to the field of depth prediction, and holds great promise for a wide range of applications in computer vision. Future research could explore the generation of accurate multi-hypotheses future depth, building upon the promising results presented in this paper. 

\bibliographystyle{IEEEtran}
\bibliography{egbib}

\begin{thebibliography}{10}
\providecommand{\url}[1]{#1}
\csname url@rmstyle\endcsname
\providecommand{\newblock}{\relax}
\providecommand{\bibinfo}[2]{#2}
\providecommand\BIBentrySTDinterwordspacing{\spaceskip=0pt\relax}
\providecommand\BIBentryALTinterwordstretchfactor{4}
\providecommand\BIBentryALTinterwordspacing{\spaceskip=\fontdimen2\font plus
\BIBentryALTinterwordstretchfactor\fontdimen3\font minus
  \fontdimen4\font\relax}
\providecommand\BIBforeignlanguage[2]{{%
\expandafter\ifx\csname l@#1\endcsname\relax
\typeout{** WARNING: IEEEtran.bst: No hyphenation pattern has been}%
\typeout{** loaded for the language `#1'. Using the pattern for}%
\typeout{** the default language instead.}%
\else
\language=\csname l@#1\endcsname
\fi
#2}}

\bibitem{Watson2021}
J.~Watson, O.~Mac~Aodha, V.~Prisacariu, G.~Brostow, and M.~Firman, ``The
  temporal opportunist: Self-supervised multi-frame monocular depth,'' in
  \emph{Proceedings of the IEEE/CVF Conference on Computer Vision and Pattern
  Recognition}, 2021, pp. 1164--1174.

\bibitem{guizilini2022multi}
V.~Guizilini, R.~Ambruș, D.~Chen, S.~Zakharov, and A.~Gaidon, ``Multi-frame
  self-supervised depth with transformers,'' in \emph{Proceedings of the
  IEEE/CVF Conference on Computer Vision and Pattern Recognition}, 2022, pp.
  160--170.

\bibitem{boulahbal2022forecasting}
H.~E. Boulahbal, A.~Voicila, and A.~I. Comport, ``Forecasting of depth and
  ego-motion with transformers and self-supervision,'' in \emph{2022 26th
  International Conference on Pattern Recognition (ICPR)}.\hskip 1em plus 0.5em
  minus 0.4em\relax IEEE, 2022, pp. 3706--3713.

\bibitem{Eigen}
D.~Eigen, C.~Puhrsch, and R.~Fergus, ``{Depth map prediction from a single
  image using a multi-scale deep network},'' in \emph{Advances in Neural
  Information Processing Systems}, vol.~3, 2014, pp. 2366--2374.

\bibitem{boulahbal2022instance}
H.~E. Boulahbal, A.~Voicila, and A.~I. Comport, ``Instance-aware multi-object
  self-supervision for monocular depth prediction,'' \emph{IEEE Robotics and
  Automation Letters}, vol.~7, no.~4, pp. 10\,962--10\,968, 2022.

\bibitem{safadoust2021self}
S.~Safadoust and F.~G{\"u}ney, ``Self-supervised monocular scene decomposition
  and depth estimation,'' in \emph{2021 International Conference on 3D Vision
  (3DV)}.\hskip 1em plus 0.5em minus 0.4em\relax IEEE, 2021, pp. 627--636.

\bibitem{lee2021attentive}
S.~Lee, F.~Rameau, F.~Pan, and I.~S. Kweon, ``Attentive and contrastive
  learning for joint depth and motion field estimation,'' in \emph{Proceedings
  of the IEEE/CVF International Conference on Computer Vision}, 2021, pp.
  4862--4871.

\bibitem{Wang2021}
L.~Wang, Y.~Wang, L.~Wang, Y.~Zhan, Y.~Wang, and H.~Lu, ``{Can Scale-Consistent
  Monocular Depth Be Learned in a Self-Supervised Scale-Invariant Manner?}''
  \emph{Proceedings of the IEEE/CVF International Conference on Computer Vision
  (ICCV)}, pp. 12\,727--12\,736, 2021.

\bibitem{Ranjan2019}
A.~Ranjan, V.~Jampani, L.~Balles, K.~Kim, D.~Sun, J.~Wulff, and M.~J. Black,
  ``{Competitive collaboration: Joint unsupervised learning of depth, camera
  motion, optical flow and motion segmentation},'' in \emph{Proceedings of the
  IEEE Computer Society Conference on Computer Vision and Pattern Recognition},
  vol. 2019-June, 2019, pp. 12\,232--12\,241.

\bibitem{Gordon2019}
A.~Gordon, H.~Li, R.~Jonschkowski, and A.~Angelova, ``{Depth from videos in the
  wild: Unsupervised monocular depth learning from unknown cameras},''
  \emph{Proceedings of the IEEE International Conference on Computer Vision},
  vol. 2019-Octob, pp. 8976--8985, 2019.

\bibitem{Godard2017}
C.~Godard, O.~{Mac Aodha}, and G.~J. Brostow, ``{Unsupervised monocular depth
  estimation with left-right consistency},'' \emph{Proceedings - 30th IEEE
  Conference on Computer Vision and Pattern Recognition, CVPR 2017}, vol.
  2017-January, pp. 6602--6611, 2017.

\bibitem{Zhou2017}
T.~Zhou, M.~Brown, N.~Snavely, and D.~G. Lowe, ``{Unsupervised learning of
  depth and ego-motion from video},'' \emph{Proceedings - 30th IEEE Conference
  on Computer Vision and Pattern Recognition, CVPR 2017}, vol. 2017-January,
  pp. 6612--6621, 2017.

\bibitem{Godard2019}
C.~Godard, O.~M. Aodha, M.~Firman, and G.~Brostow, ``{Digging into
  self-supervised monocular depth estimation},'' \emph{Proceedings of the IEEE
  International Conference on Computer Vision}, vol. 2019-October, no.~1, pp.
  3827--3837, 2019.

\bibitem{Johnston2020}
A.~Johnston and G.~Carneiro, ``{Self-supervised monocular trained depth
  estimation using self-attention and discrete disparity volume},''
  \emph{Proceedings of the IEEE Computer Society Conference on Computer Vision
  and Pattern Recognition}, pp. 4755--4764, 2020.

\bibitem{Rares2020}
V.~G. Rares, Ambrus, S.~Pillai, A.~Raventos, and A.~Gaidon, ``{3D packing for
  self-supervised monocular depth estimation},'' \emph{Proceedings of the IEEE
  Computer Society Conference on Computer Vision and Pattern Recognition}, pp.
  2482--2491, 2020.

\bibitem{9864127}
S.-J. Hwang, S.-J. Park, J.-H. Baek, and B.~Kim, ``Self-supervised monocular
  depth estimation using hybrid transformer encoder,'' \emph{IEEE Sensors
  Journal}, vol.~22, no.~19, pp. 18\,762--18\,770, 2022.

\bibitem{Mahjourian2017}
R.~Mahjourian, M.~Wicke, and A.~Angelova, ``{Geometry-based next frame
  prediction from monocular video},'' in \emph{IEEE Intelligent Vehicles
  Symposium, Proceedings}, 2017, pp. 1700--1707.

\bibitem{Qi2019}
X.~Qi, Z.~Liu, Q.~Chen, and J.~Jia, ``{3D motion decomposition for RGBD future
  dynamic scene synthesis},'' in \emph{Proceedings of the IEEE Computer Society
  Conference on Computer Vision and Pattern Recognition}, vol. 2019-June, 2019,
  pp. 7665--7674.

\bibitem{Hu2020}
A.~Hu, F.~Cotter, N.~Mohan, C.~Gurau, and A.~Kendall, ``{Probabilistic Future
  Prediction for Video Scene Understanding},'' in \emph{Lecture Notes in
  Computer Science (including subseries Lecture Notes in Artificial
  Intelligence and Lecture Notes in Bioinformatics)}, vol. 12361 LNCS, 2020,
  pp. 767--785.

\bibitem{Geiger2012CVPR}
A.~Geiger, P.~Lenz, and R.~Urtasun, ``Are we ready for autonomous driving? the
  kitti vision benchmark suite,'' in \emph{Conference on Computer Vision and
  Pattern Recognition (CVPR)}, 2012.

\bibitem{Shu2020}
C.~Shu, K.~Yu, Z.~Duan, and K.~Yang, ``{Feature-Metric Loss for Self-supervised
  Learning of Depth and Egomotion},'' \emph{Lecture Notes in Computer Science
  (including subseries Lecture Notes in Artificial Intelligence and Lecture
  Notes in Bioinformatics)}, vol. 12364 LNCS, pp. 572--588, 2020.

\bibitem{Bian2019}
J.~Bian, Z.~Li, N.~Wang, H.~Zhan, C.~Shen, M.-M. Cheng, and I.~Reid,
  ``Unsupervised scale-consistent depth and ego-motion learning from monocular
  video,'' \emph{Advances in neural information processing systems}, vol.~32,
  pp. 35--45, 2019.

\bibitem{Chen2019}
Y.~Chen, C.~Schmid, and C.~Sminchisescu, ``{Self-supervised learning with
  geometric constraints in monocular video: Connecting flow, depth, and
  camera},'' \emph{Proceedings of the IEEE International Conference on Computer
  Vision}, vol. 2019-October, pp. 7062--7071, 2019.

\bibitem{chen2019self}
------, ``Self-supervised learning with geometric constraints in monocular
  video: Connecting flow, depth, and camera,'' in \emph{Proceedings of the
  IEEE/CVF International Conference on Computer Vision}, 2019, pp. 7063--7072.

\bibitem{Klingner2020}
M.~Klingner, J.~A. Term{\"{o}}hlen, J.~Mikolajczyk, and T.~Fingscheidt,
  ``{Self-supervised Monocular Depth Estimation: Solving the Dynamic Object
  Problem by Semantic Guidance},'' \emph{Lecture Notes in Computer Science
  (including subseries Lecture Notes in Artificial Intelligence and Lecture
  Notes in Bioinformatics)}, vol. 12365 LNCS, pp. 582--600, 2020.

\bibitem{Vijayanarasimhan}
S.~Vijayanarasimhan, S.~Ricco, C.~Schmid, R.~Sukthankar, and K.~Fragkiadaki,
  ``Sfm-net: Learning of structure and motion from video,'' \emph{arXiv
  preprint arXiv:1704.07804}, 2017.

\bibitem{Lee2019}
S.~Lee, S.~Im, S.~Lin, and I.~S. Kweon, ``{Learning Residual Flow as Dynamic
  Motion from Stereo Videos},'' \emph{IEEE International Conference on
  Intelligent Robots and Systems}, pp. 1180--1186, 2019.

\bibitem{johnston2020self}
A.~Johnston and G.~Carneiro, ``Self-supervised monocular trained depth
  estimation using self-attention and discrete disparity volume,'' in
  \emph{Proceedings of the ieee/cvf conference on computer vision and pattern
  recognition}, 2020, pp. 4756--4765.

\bibitem{wimbauer2021monorec}
F.~Wimbauer, N.~Yang, L.~Von~Stumberg, N.~Zeller, and D.~Cremers, ``Monorec:
  Semi-supervised dense reconstruction in dynamic environments from a single
  moving camera,'' in \emph{Proceedings of the IEEE/CVF Conference on Computer
  Vision and Pattern Recognition}, 2021, pp. 6112--6122.

\bibitem{jaderberg2015spatial}
M.~Jaderberg, K.~Simonyan, A.~Zisserman, and K.~Kavukcuoglu, ``Spatial
  transformer networks,'' in \emph{Proceedings of the 28th International
  Conference on Neural Information Processing Systems-Volume 2}, 2015, pp.
  2017--2025.

\bibitem{liu2021swin}
Z.~Liu, Y.~Lin, Y.~Cao, H.~Hu, Y.~Wei, Z.~Zhang, S.~Lin, and B.~Guo, ``Swin
  transformer: Hierarchical vision transformer using shifted windows,'' in
  \emph{Proceedings of the IEEE/CVF international conference on computer
  vision}, 2021, pp. 10\,012--10\,022.

\bibitem{deng2009imagenet}
J.~Deng, W.~Dong, R.~Socher, L.-J. Li, K.~Li, and L.~Fei-Fei, ``Imagenet: A
  large-scale hierarchical image database,'' in \emph{2009 IEEE conference on
  computer vision and pattern recognition}.\hskip 1em plus 0.5em minus
  0.4em\relax Ieee, 2009, pp. 248--255.

\bibitem{wang2004image}
Z.~Wang, A.~C. Bovik, H.~R. Sheikh, and E.~P. Simoncelli, ``Image quality
  assessment: from error visibility to structural similarity,'' \emph{IEEE
  transactions on image processing}, vol.~13, no.~4, pp. 600--612, 2004.

\bibitem{cordts2016cityscapes}
M.~Cordts, M.~Omran, S.~Ramos, T.~Rehfeld, M.~Enzweiler, R.~Benenson,
  U.~Franke, S.~Roth, and B.~Schiele, ``The cityscapes dataset for semantic
  urban scene understanding,'' in \emph{Proceedings of the IEEE conference on
  computer vision and pattern recognition}, 2016, pp. 3213--3223.

\bibitem{RobotCarDatasetIJRR}
W.~Maddern, G.~Pascoe, C.~Linegar, and P.~Newman, ``{1 Year, 1000km: The Oxford
  RobotCar Dataset},'' \emph{The International Journal of Robotics Research
  (IJRR)}, vol.~36, no.~1, pp. 3--15, 2017.

\bibitem{4359315}
H.~Hirschmuller, ``Stereo processing by semiglobal matching and mutual
  information,'' \emph{IEEE Transactions on Pattern Analysis and Machine
  Intelligence}, vol.~30, no.~2, pp. 328--341, 2008.

\bibitem{kingma2014adam}
D.~P. Kingma and J.~Ba, ``Adam: A method for stochastic optimization,''
  \emph{arXiv preprint arXiv:1412.6980}, 2014.

\bibitem{yang2018unsupervised}
Z.~Yang, P.~Wang, W.~Xu, L.~Zhao, and R.~Nevatia, ``Unsupervised learning of
  geometry from videos with edge-aware depth-normal consistency,'' in
  \emph{Thirty-Second AAAI conference on artificial intelligence}, 2018.

\bibitem{Yin2018}
Z.~Yin and J.~Shi, ``{GeoNet: Unsupervised Learning of Dense Depth, Optical
  Flow and Camera Pose},'' \emph{Proceedings of the IEEE Computer Society
  Conference on Computer Vision and Pattern Recognition}, pp. 1983--1992, 2018.

\end{thebibliography}

\end{document}